\documentclass{article}
\usepackage{spconf,amsmath,graphicx}
\usepackage{times}
\usepackage{epsfig}
\usepackage{graphicx}
\usepackage{amsmath}
\usepackage{amssymb}
\usepackage{subcaption}
\usepackage{multicol}
\usepackage{multirow}
\usepackage{amsmath}
\usepackage{amssymb}
\usepackage{xcolor}
\usepackage{enumerate}
\usepackage{caption}

\def\eg{\emph{e.g.} }

\def\etal{\emph{et al.} }
\newcommand{\ttilde}{$\sim$}%{\textasciitilde}

\title{Improving Object Detection and Attribute Recognition by \\Feature Entanglement Reduction}
\name{Zhaoheng Zheng, Arka Sadhu and Ram Nevatia\sthanks{This material is based upon work supported by the Defense Advanced Research Projects Agency (DARPA) under Agreement No. HR00111990059.}}
\address{University of Southern California \\
Los Angeles, USA}
\begin{document}
%\ninept
%
\maketitle
\begin{abstract}
We explore object detection with two attributes: color and material. The task aims to simultaneously detect objects and infer their color and material. A straight-forward approach is to add attribute heads at the very end of a usual object detection pipeline. However, we observe that the two goals are in conflict: Object detection should be attribute-independent and attributes be largely object-independent. Features computed by a standard detection network entangle the category and attribute features; we disentangle them by the use of a two-stream model where the category and attribute features are computed independently but  the classification heads share Regions of Interest (RoIs). Compared with a traditional single-stream model, our model shows significant improvements over VG-20,  a subset of Visual Genome, on both supervised and attribute transfer tasks.
\end{abstract}
\section{Introduction}

\label{sec:intro}

Object detection has seen tremendous progress through deep neural networks \cite{ren2015faster,he2017mask, cai2018cascade,pang2019libra,Nie_2019_ICCV,Duan_2019_ICCV} and availability of large scale datasets such as MS-COCO \cite{lin2014microsoft} and Visual Genome \cite{krishnavisualgenome}. 
In addition to objects, attributes are useful in distinguishing among members of the same category.  
While attribute recognition is a classic topic when applied to homogeneous patches, the task is much more complex when applied to an entire object. 
Recently, joint prediction of objects and attributes has been explored under scene-graph generation \cite{Liang2017DeepVR}, symbolic VQA \cite{yi2018neural}, dense captioning \cite{densecap} and image captioning \cite{anderson2018bottom}. 
In particular, the model used in \cite{anderson2018bottom} has been widely adopted as the feature extractor for VQA tasks, and as such forms a competitive baseline in our experiments. 
However, prior work does not evaluate performance on novel object-attribute pairs; in this paper, we explore the usual object-attribute detection problem and  extension to recognition of novel object-attribute pairs.

%While attribute classification is a traditional topic when applied to homogeneous patches, the task is much more complex when applied to an entire object. It is also not feasible to learn all combinations of attributes and categories due to their large numbers. A key observation of our work is that the feature requirements of the two tasks are in conflict: category classification should be invariant to attributes and attribute classification should be invariant to category. We present an approach to resolving this conflict and make possible recognition of novel attribute, category combinations.

\begin{figure}[t]
\centering
\begin{subfigure}[t]{.15\textwidth}
  \centering
  \includegraphics[height=\textwidth, width=\textwidth]{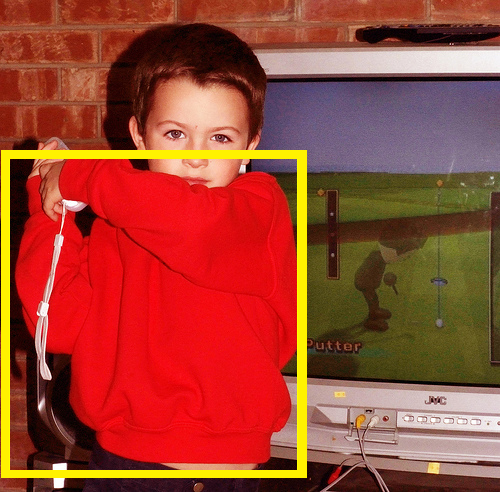}  
  \caption{shirt \newline\centering \textit{red}}
  \label{fig:red-shirt}
\end{subfigure}
\begin{subfigure}[t]{.15\textwidth}
  \centering
  \includegraphics[height=\textwidth, width=\textwidth]{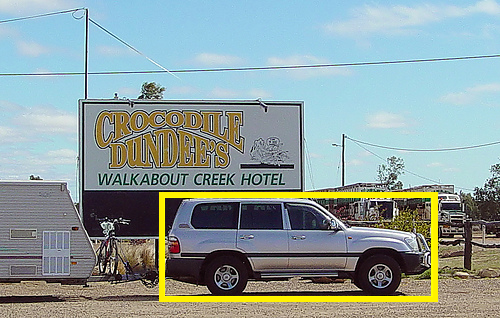}
  \caption{\centering car \newline \centering\textit{unlabelled}}
  \label{fig:car}
\end{subfigure}
\begin{subfigure}[t]{.15\textwidth}
  \centering
  \includegraphics[height=\textwidth, width=\textwidth]{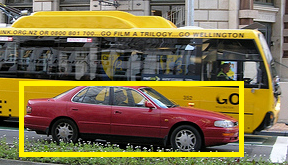}
  \caption{\centering car \newline \centering \textit{red}}
  \label{fig:yellow-car}
\end{subfigure}
\caption{Attribute Transfer: During training phase, some categories \eg{(a)} come with attribute labels, while some \eg{(b)} only have object class labelled. Models need to recognize attributes on such \textit{unlabelled} categories, \eg{(c)} during testing.}
\label{fig:task-description}
\end{figure}

In Fig. \ref{fig:task-description}, we show some examples of objects with color attributes. Note that the ``red car'' can be distinguished from a ``silver car'' based on color . We note that the property of color is not specific to the car. 
Unlike naming color on patches \cite{han2019grounding,winn2018lighter}, recognizing the color of an object is more challenging. 
Typical objects are not of a single, uniform hue 
% and there are also 
with further 
variations due to changes in surface orientation, illumination, reflections, shadows, and highlights. 
The material composition may also not be uniform; for example, a car has both metal and glass components. 
One other difficulty is created with the use of rectangular bounding boxes for object proposals which mix background pixels with object pixels. 
We do not aim to separate these influences; instead, as in object classification, we aim to learn from examples where the variations are accounted for in a holistic feature vector.  
%It is important that an object, attribute recognizer be able to recognize novel combinations of such entities and not expect all such combinations to have been seen in the training data. 

A further challenge is that common detection datasets do not come with attribute annotations; even in those, such as Visual Genome \cite{krishnavisualgenome}, that do provide attributes, a large proportion of objects is not attribute annotated. Additionally, as shown in Fig. \ref{fig:task-description}, it is not reasonable to expect training data to contain all possible attribute-category pairs; a desirable model needs to recognize novel attribute-category pairs not encountered in training, we name this task as one of attribute transfer.

There is an inherent conflict between the feature requirements of category and attribute classification tasks: the former aims to be largely attribute-invariant 
%(e.g. we want a car of any color to be still classified as a car) 
and the latter to be largely invariant to category class 
%(e.g. a red car and a red rose are both red though possibly different shades of hues and saturation in detail). 
Simply attribute classification heads to the end of a two-stage detection pipeline (for instance, Faster R-CNN \cite{ren2015faster}) entangles the features for two conflicting needs, weighing on the performance of both object detection and attribute prediction.

To eliminate potential entanglement in feature space, we separate the feature extraction into two streams.
%We argue that it is better if the attribute and category classifiers have separate feature extraction streams to avoid the inherent conflict. 
More specifically, in our proposed model, category classifier and attribute classifier are fed with separate features from two independent convolutional backbones, while region proposals are shared. 
%{\color{black} We do also explore limited amount of interaction between object categories and attributes to model their dependencies (in Sec. \ref{sec:ts-arch}). We use the same stream for material and color attributes as we do not expect them to be in conflict as both compute averaged properties and are not dependent on the object shape; this expectation is validated by our experiments on separate and joint  attribute recognition( in Sec. \ref{sec:result}).}

We evaluate the accuracy of single-stream and two-stream variants on VG-20, which we construct from the Visual Genome \cite{krishnavisualgenome} dataset.
% We also construct splits from these datasets to investigate attribute transfer.
We further construct novel splits from these datasets and investigate the ability of the models to transfer attributes. 
Our experiments show that in a single-stream architecture, incorporating attribute heads results in a significant drop in object detection mAP whereas there is little or no loss in the two-stream variants under novel attribute-category combinations and that the two-stream architecture achieves higher attribute transfer accuracy.

Our contributions are:
(i) we eliminate the feature entanglement and resolve the internal conflict between object detection and attribute classification through the two-stream design;
(ii) VG-20, a new subset of Visual Genome and splits in this dataset for evaluating attribute inference and transfer performance;
% and 
(iii) demonstration of significant improvements over baselines for attribute inference and transfer tasks.

\section{Method}

% In this section, we first revisit the R-CNN framework in Sec. \ref{sec:background} and introduce our two-stream architecture in Sec. \ref{sec:ts-arch}.
\textbf{R-CNN Detection Structure:} Recent detection structures in the R-CNN family are
composed of four parts: a deep convolutional backbones like ResNet \cite{he2016deep} and VGG \cite{Simonyan15} , the Region Proposal Network (RPN) \cite{ren2015faster}, a feature extractor and a classification module. Specifically, the convolutional backbone processes image-level
features from input images, and the RPN, takes features to generate proposals, or in other words, RoIs. The RoI feature extractor extracts features for these regions of interest via RoI Pooling operations. The classification module uses these RoI features to classify and regress the bounding box. In our case, we additionally have an attribute head for attribute classification. 

\textbf{Attribute Recognition with Object Embedding:} Anderson \etal \cite{anderson2018bottom} introduced an additional substructure designed for attribute recognition. 
% Such a substructure consists of two layers: an embedding layer and a linear layer. 
An embedding layer maps object labels into features, which are concatenated with RoI features, followed by a linear layer and finally fed to the classification heads for prediction. 
% Note that the embedding layer takes object labels during training and top object predictions when testing.
Such a structure brings object information to the attribute classification. But the conflict between object detection and attribute recognition remains.

\begin{figure}[t]
    \centering
    \includegraphics[width=\linewidth]{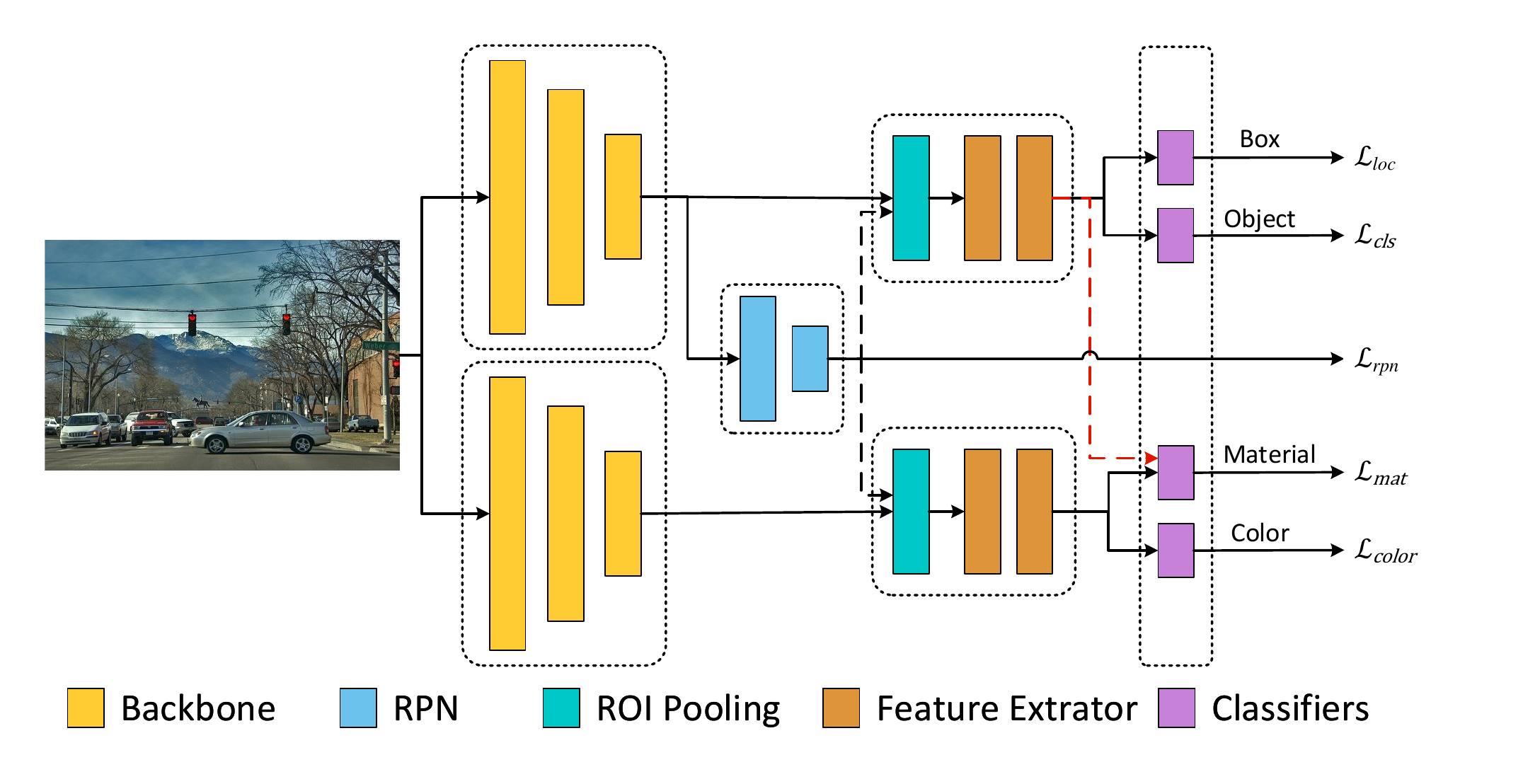}
    \caption{Our two-stream architecture: Note that solid arrows are active in both inference and back propagation while dotted arrows are only active in inference.}
    \label{fig:pipeline}
\end{figure}

\textbf{Two-Stream Architecture:}
The proposed architecture follows the R-CNN pipeline, with the backbone and RoI feature extractor divided into two independent streams, as shown in Fig. \ref{fig:pipeline}. 
The object head and box head are kept in the object stream while the attribute stream makes attribute predictions. {\color{black} We choose to use the same stream for both color and material attributes based on similar requirements of feature extraction. }
On top of the R-CNN pipeline, we add an attribute feature extractor which uses the RoI from RPN.
The RPN is integrated into the object stream and takes features from the object backbone to generate region proposals. 
Region proposals are shared by both streams because attributes are properties associated with certain objects and computing proposals solely for attributes would be meaningless.
% and it is meaningless to compute proposals solely for attributes. 

\textbf{Cross Link: } In the ordinary two-stream design, the object stream (the top side) and the attribute stream (the bottom side) make predictions from separate features computed by independent feature extractors. 
But unlike objects and color, objects and material are highly correlated. While some objects appear in some colors only, most man-made objects can appear in a variety of colors whereas the material property is much more constrained. %For instance, a cooking pan can be any color while it can’t be made of plastic. 
%For instance, a door is very likely to be wooden while most windows are made of glass. 
To leverage such correlation, we add a cross link, the red dotted arrow in Fig. \ref{fig:pipeline}, from the object stream to the attribute stream. Features are concatenated before the prediction layer. Furthermore, the gradient from the material head to the object stream is blocked so that utilizing object features for attributes will not impair object detection.

\textbf{Objectives:}
\label{sec:label}
The overall loss function $\mathcal{L}$ is the sum of four components $\mathcal{L}_{rpn}, \mathcal{L}_{loc}, \mathcal{L}_{cls}, \mathcal{L}_{attr}$, that is,
% \begin{equation}
    $\mathcal{L} = \mathcal{L}_{rpn} + \mathcal{L}_{loc} + \mathcal{L}_{cls} + \mathcal{L}_{attr}$.
% \end{equation}
In terms of $\mathcal{L}_{loc}$ and $\mathcal{L}_{cls}$, we follow the same objective function proposed in \cite{girshick2015fast}. As for $\mathcal{L}_{rpn}$, the loss function of RPN, we follow the one defined in \cite{ren2015faster}. 
And $\mathcal{L}_{attr} = \mathcal{L}_{color} + \mathcal{L}_{mat}$
% And $\mathcal{L}_{attr}$ is the sum of $\mathcal{L}_{color}$ and $\mathcal{L}_{mat}$: 
% \begin{equation}
%     \mathcal{L}_{attr} = \mathcal{L}_{color} + \mathcal{L}_{mat},
% \end{equation}
Here,
\begin{align}
    &\mathcal{L}_{color} = H(\sigma(\textbf{z}_{color}),\textbf{y}_{color}), \label{eqn:SCE-1}\\ 
    &\mathcal{L}_{mat} = H(\sigma(\textbf{z}_{mat}), \textbf{y}_{mat}).
    \label{eqn:SCE-2}
\end{align}
Note that $H$ is the cross-entropy loss, $\sigma$ refers to the softmax function, and $\textbf{z}, \textbf{y}$ are inference scores and labels respectively. We name $L_{attr}$ as Separated Cross-Entropy loss (SCE) given that it is the sum of two independent cross-entropy functions.

\section{Experiments}
\label{sec:exp}
We introduce our data preparation in Sec. \ref{sec:data} and detail our experimental setup in Sec. \ref{sec:setting}, followed by quantitative results in Sec. \ref{sec:result} and qualitative visualizations in  Sec. \ref{ss:vis}.

\subsection{Data Preparation}
\label{sec:data}

To evaluate the performance of our approach, we construct a subset of Visual Genome \cite{krishnavisualgenome}; specifically, we adopt the split and reorganized scene graphs created by Hudson \etal \cite{hudson2018gqa}. The Visual Genome dataset consists of 108k images along with over 1,600 object categories and around 400 attribute annotations associated with objects. However,  many categories in the dataset overlap with other categories (for example, ``man'' and ``person'' are labeled as two different categories) and it also suffers from a long-tailed attribute distribution. Therefore, we pick 12 most descriptive colors and 4 most common materials from the dataset. Regarding object categories, we select 20 categories that have sufficient attribute annotations for our task. Thus we call our dataset as VG-20. In total, we have 180k samples for training and 25k for testing, with around one-third of them possess attribute annotations. Note that each bounding box is counted as one object sample and some bounding boxes do not have associated attribute annotations; we preserve these as they are useful in both training and evaluating object detectors. 

\begin{table}[t]
\centering
\resizebox{\linewidth}{!}{%
\begin{tabular}{l|ccc}
\hline
\multirow{2}{*}{Model}                & Object mAP  & Color Recall & Material Recall \\
& @.5 & @.5 & @.5\\ \hline \hline
PA + SCE   & 24.95  & 67.77  & 56.98   \\
PA + UCE   & 25.35  & 68.74 & 56.01   \\
Single-Stream (SS)      & 25.13  & 68.59 & 61.22   \\
SS Detection Only & 38.18  & - & -\\
Two-Stream (TS)   & 38.17 & 72.40  & 63.83
      \\
TS + Cross Link & \textbf{38.30} & \textbf{73.11} & \textbf{65.39} \\
TS + LFE & 28.37  & 72.72 & 63.50  \\
%TS w/o Material & 38.37 (41.46) & 72.47 (73.97) & - &  82.77 & 75.81 & - \\
%TS + CL w/o Color & 38.61 (41.69) & - & 63.03 (67.08) & 82.22 & - & 79.87 \\ 

\hline
\end{tabular}%
}
\caption{Results of Supervised Object Detection and Attribute Prediction: PA refers to the detection model used in \cite{anderson2018bottom}. LFE refers to the variation with Late Fusion Entanglement.}
\label{tab:gqa-sup}
\end{table}

\begin{table}[t]
\centering
\resizebox{\linewidth}{!}{%
\begin{tabular}{l|l|ccc}
\hline
\multirow{2}{*} & \multirow{2}{*}{Model}                & Object mAP  & Color Recall  & Material Recall \\& & @.5  & @.5  & @.5 \\ \hline \hline
\multirow{5}{*}{Target} & PA + SCE & 25.09 & 50.89 & 47.85       \\
                        & PA + UCE & 24.13 & 52.75 & 46.40  \\ 
                        & Single-Stream    & 24.87  & 48.39 & 49.92   \\ 
                        & Two-Stream (TS)     & 38.11  & 54.98  & 49.27 \\ 
                        & TS + Cross Link  & \textbf{38.39} & \textbf{61.01} & \textbf{52.28}  \\
                        & TS + LFE  & 27.39 & 46.55 & 48.93 \\
                        \hline \hline
\multirow{5}{*}{Reference} & PA + SCE  & 23.27 & 67.26 & 59.70 \\
                        & PA + UCE & 22.32 & 66.41 & 58.62  \\
                        & Single-Stream & 22.76  & 67.39 & 61.48  \\
                        & Two-Stream (TS)   & 37.67 & 69.43 & 62.67 \\
                        & TS + Cross Link  & \textbf{38.26}  & \textbf{71.73}  & \textbf{66.14} \\ 
                        & TS + LFE  & 26.11 & 68.91 & 63.25\\
                         \hline
\end{tabular}%
}
\caption{Results of Attribute Transfer: We use metrics object mAP, color recall, material recall, as defined in Sec. \ref{sec:setting}. SCE and UCE are loss functions defined in Eqn. \ref{eqn:SCE-1}, \ref{eqn:SCE-2} and Eqn. \ref{eqn:UCE}. And PA refers to the detection model in \cite{anderson2018bottom}.}
\label{tab:gqa-zs}
\end{table}

\subsection{Experimental Setup}
\label{sec:setting}         
% \textbf{Experiment Settings:} 
We explore two settings w.r.t attribute annotations: 
\begin{itemize}
\itemsep0em  
    \item \textbf{Supervised}: all attribute annotations are available during training phase.
    \item \textbf{Attribute Transfer}: objects are divided into two groups by their object labels, reference categories $X_{ref}$ and target categories $X_{tgt}$. 
    During training, objects in $X_{ref}$ keep their attribute annotations while those in $X_{tgt}$ do not have access to the attribute annotations. 
    That is, the model needs to transfer attributes from $X_{ref}$ to $X_{tgt}$ which brings additional complexity.
\end{itemize}
% (i) \textbf{supervised}: all attribute annotations are available during training phase
% (ii) \textbf{attribute transfer}: objects are divided into two groups by their object labels, reference categories $X_{ref}$ and target categories $X_{tgt}$. 
% During training, objects in $X_{ref}$ keep their attribute annotations while those in $X_{tgt}$ do not have access to the attribute annotations. 
% That is, the model needs to transfer attributes from $X_{ref}$ to $X_{tgt}$ and brings additional complexity to the task.
% In this scenario, a model is required to learn attributes from $X_{ref}$ and transfer the knowledge to recognize the attributes of $X_{tgt}$, which conveys more challenges to the task.

\iffalse
\begin{table}[t]
    \centering
    \begin{tabular}{c|c|c|c|c|c}
    \hline \hline
         \multirow{2}{*}{$X_A$} & car & hat & bird & bottle & bear  \\
         & tree & door & chair & table & fence \\ \hline
         \multirow{2}{*}{$X_B$}& pole & dog & truck & window & bench \\
         & cup & shirt & cat & post & cow \\ \hline
    \end{tabular}
    \caption{Group Division in Attribute Transfer on VG-20}
    \label{tab:split}
\end{table}
\fi

For fair evaluation of attribute transfer over all categories, we divide the objects into two groups $X_A$ and $X_B$, which satisfy following properties: $X_A \cap X_B = \varnothing ,\, X_A \cup X_B = X_{all}$ and keep $|X_A| {=} |X_B| {=} 10$.
We let $X_{ref} = X_A,\, X_{tgt}=X_B$ in one run and vice versa in the other. 
Quantitative numbers are averaged over those two runs. 
%Our group division on VG-20 is shown in \ref{tab:split}. 
% Such division is made according to the distribution of attributes so that each attributes have sufficient number of samples in each group.

\textbf{Evaluation Metrics:} 
For object detection, we adopt the commonly used mean Average Precision (mAP@0.5). %We also note that mAP is slightly unreliable for VG due to its sparse annotations where some correct detections could be evaluated as false positives. 
% since VG is sparsely annotated, mAP is somewhat unreliable as some correct detections are evaluated as false alarms due to missing ground truth bounding boxes.
Furthermore, to measure both detection and recognition performances simultaneously, we define ``attribute recall" (attribute could be color or material) as the ratio of objects whose bounding boxes and attributes are detected and recognized by the model to all objects with valid attribute annotations.

\begin{figure*}[ht!]
\centering
\includegraphics[width=.88\textwidth]{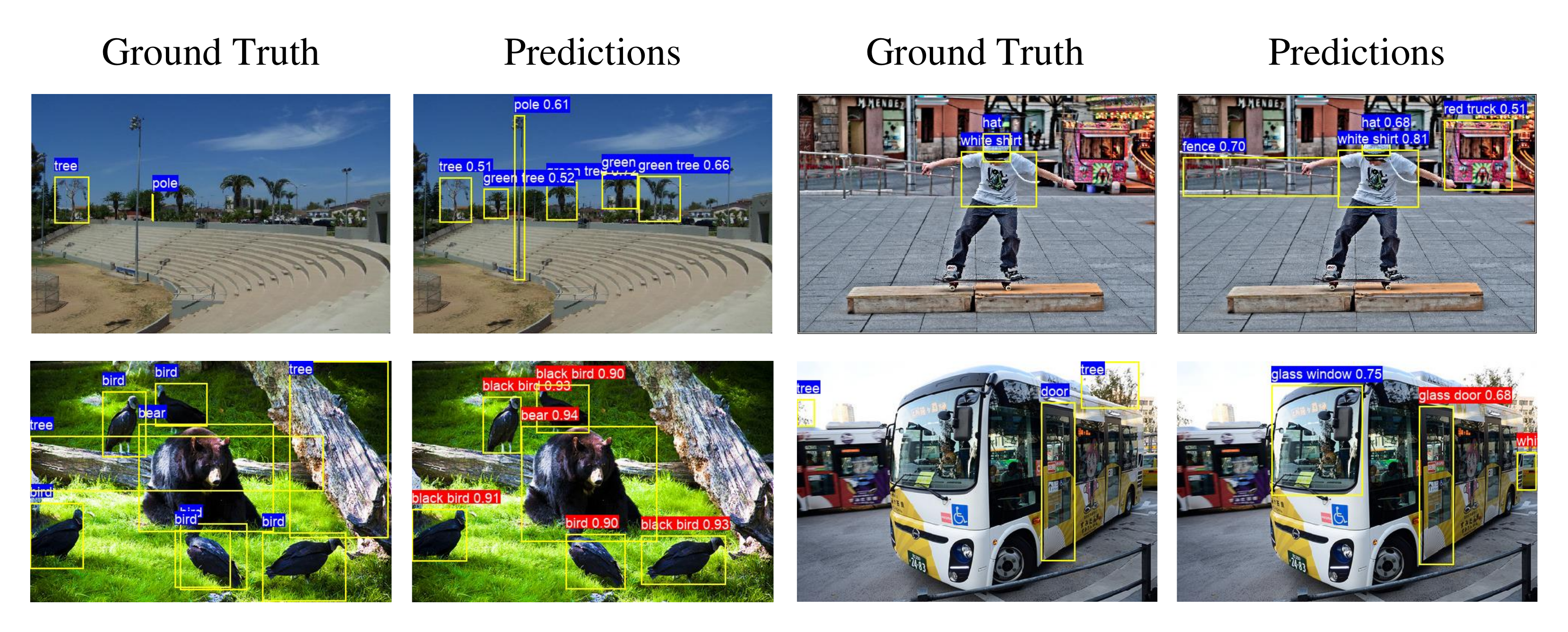}
\caption{\color{black} Visualized detection results for VG-20 supervised (row 1), VG-20 attribute transfer (row 2). Object predictions colored in blue belong to reference set while those colored in red belong to target set in the attribute transfer setting. 
%Note that some correct detections, for example the fence in first image of row2,  are not annotated in the ground truth and get counted as false alarms in quantitative evaluation
}
\label{fig:vis}
\end{figure*}

\textbf{Baselines:} 
We compare our model against two baseline approaches and one variation of our design.
\begin{itemize}
\itemsep0em  
\item \textit{Single-Stream}: A single-stream version of our model.
\item \textit{Peter Anderson Model (PA)}: The R-CNN-like structure proposed in \cite{anderson2018bottom}. For fair comparison, we integrate an FPN to this model and retrain it with our data splits. 
The original PA model uses Unified Cross-Entropy loss:
\begin{equation}
    \mathcal{L}_{attr} = H(\sigma(\textbf{z}), \textbf{y}),
    \label{eqn:UCE}
\end{equation}
where each color and material is treated as an attribute. 
We compare with two variants of PA, one trained with Unified Cross-Entropy (UCE) and the other with Separated Cross-Entropy (SCE), referred as PA + UCE and PA + SCE respectively.
\item \textit{Late Feature Entanglement (LFE)}: A variation of our two-stream model where features from both streams are explicitly entangled. More specifically, RoI features from both streams are concatenated  before classification so that all classifiers share identical features.
\end{itemize}

\textbf{Implementation Details:} 
We adopt ResNet-101 \cite{he2016deep} as our backbones in both streams and the design of RPN follows \cite{ren2015faster}. 
We build the feature extractors following Feature Pyramid Networks (FPN) in \cite{lin2017feature}. 
%Besides, each classification head consists of one single-layer MLP. 
%Our model is implemented with PyTorch \cite{paszke2019pytorch}. 
During training, both streams of our model are initialized with the pre-trained weights from MSCOCO \cite{lin2014microsoft}. 
The model is trained by Adam \cite{kingma205adam} Optimizer with a learning rate of $5e-5$. The batch size is set to $12$. 
%The random seed is selected by the default built-in mechanism inside PyTorch. Our code partially comes from \textit{maskrcnn-benchmark}\footnote{https://github.com/facebookresearch/maskrcnn-benchmark} and we follow their hyper-parameter settings.}

\subsection{Quantitative Results}
\label{sec:result}

We report results for VG-20 in both supervised and attribute transfer setting.

\begin{enumerate}[(a)]

\item \textbf{Supervised:} As seen in Table \ref{tab:gqa-sup} for VG-20, compared with the detection only model, the single-stream detection plus attribute inference model brings down the object mAP by more than 10\%. 
The two-stream variants do not exhibit this drop and also show a \ttilde 3\% boost on color recall and an \ttilde 2\% improvement on material recall. We also compare with our implementation of \cite{anderson2018bottom} with two different cross-entropy (PA + SCE and PA + UCE) 
which give similar results as single-stream models.
%Under perfect detection, our models also show performance gain on object accuracy and color accuracy.

Furthermore, the late-stage feature entanglement does not show improvements in attribute recognition and even impairs the performance of object detection, dragging down object mAP by \ttilde 10\%. By comparing \textit{Single-Stream} with \textit{TS + LFE}, we demonstrate that, even though both object detection and attribute recognition benefit from the increased number of parameters, the feature entanglement between object stream and attribute stream leads to a significant deterioration on object-related performances.

\item \textbf{Attribute Transfer:} Results on VG, shown in Tab. \ref{tab:gqa-zs}, also show noticeable improvements. 
In color domain, our models increase the performance by more than 10\% on color recall.
Finally, results on reference set are consistent with those in supervised setting as expected.

\textbf{Effectiveness of the Cross Link:} As shown in Tab. \ref{tab:gqa-zs}, the cross link improves the performance of our two-stream model especially in transferring attributes. The link improves the color recall and material recall in target categories by around 6\% and 3\%, respectively. Such results reflect that with less supervision, the cross link enables the attribute stream to learn from the object stream, resulting in the gain in attribute transfer.
\end{enumerate}

\iffalse
\begin{table}[t]
\centering
\resizebox{\linewidth}{!}{%
\begin{tabular}{l|cc|cc}
\hline
\multirow{2}{*}{Model}                 & Color Recall & Material Recall & \multirow{2}{*}{Color Acc.} & \multirow{2}{*}{Material Acc.}\\ 
& @.5 (.3) & @.5 (.3) & \\ \hline \hline
Color + Material & 73.11 (75.00) & 65.39 (69.84) & 77.97  & 79.82\\
Color Only &  72.47 (73.97) & - & 75.81 & - \\
Material Only &  - & 63.03 (67.08) &  - & 79.87 \\ 
\hline
\end{tabular}%
}
\caption{ Results  of Supervised  Object Detection and Attribute Prediction on VG-20 under Different Attribute Targets.}
\label{tab:gqa-diff}
\end{table}

\item \textbf{Relationship between Color and Material Classification:} Results in Tab. \ref{tab:gqa-diff} show that the performance of recognizing colors and materials simultaneously is comparable with recognizing them separately. As expected, these two attributes are not in conflict and can be accommodated in a single stream.
\fi

\subsection{Visualization}
\label{ss:vis}
We visualize detection results of our two-stream model in Fig. \ref{fig:vis} (only objects with confidence $\ge 0.5$ are shown).

\textbf{Supervised}: Predictions are shown in first row.
We note that the ground-truth annotations in VG are sparse (i) only some objects are annotated with bounding boxes (ii) even among objects with bounding boxes, only some are annotated with their color and material attributes. 
Though some objects are not annotated in the ground truth, our model provides reasonably dense predictions of objects and attributes.
% Second example in row-2 shows a failure case, where the edges between objects are too fine-grained to be distinguished by the object detector. 

\iffalse
\textbf{CLEVR-Ref+ Attribute Transfer:} Results are shown in row 3 and 4.
The object categories for which no attribute labels are provided in the training data are marked in ``red''.
Clearly, our model is able to transfer both color and material information from the reference categories to ``spheres''.
\fi

\textbf{Attribute Transfer:} 
Examples in the second row show that our model can transfer attribute in real-world images.
The colors of animals and materials of doors are well transferred. 
%The example at the right-bottom represents a failure case where the tree is recognized to be made of glass.
%Such failure may due to the mixture of other objects in the scene.

\section{Conclusion}
\label{sec:con}
We explore the task of jointly detecting objects and predicting their attributes.
We show that naively attaching attribute heads to an R-CNN structure and jointly training object category and attribute leads to a significant drop in object detection performance due to feature entanglement. So we eliminate such feature entanglement via a two-stream pipeline with separate networks. We validate our approach on a subset of Visual Genome, VG-20. Experiments show that our method can effectively improve the performance on both tasks.

\bibliographystyle{IEEEbib}
\bibliography{refs}

\begin{thebibliography}{10}

\bibitem{ren2015faster}
Shaoqing Ren, Kaiming He, Ross Girshick, and Jian Sun,
\newblock ``Faster r-cnn: Towards real-time object detection with region
  proposal networks,''
\newblock in {\em Advances in neural information processing systems}, 2015, pp.
  91--99.

\bibitem{he2017mask}
Kaiming He, Georgia Gkioxari, Piotr Doll{\'a}r, and Ross Girshick,
\newblock ``Mask r-cnn,''
\newblock in {\em Proceedings of the IEEE international conference on computer
  vision}, 2017, pp. 2961--2969.

\bibitem{cai2018cascade}
Zhaowei Cai and Nuno Vasconcelos,
\newblock ``Cascade r-cnn: Delving into high quality object detection,''
\newblock in {\em Proceedings of the IEEE conference on computer vision and
  pattern recognition}, 2018, pp. 6154--6162.

\bibitem{pang2019libra}
Jiangmiao Pang, Kai Chen, Jianping Shi, Huajun Feng, Wanli Ouyang, and Dahua
  Lin,
\newblock ``Libra r-cnn: Towards balanced learning for object detection,''
\newblock in {\em Proceedings of the IEEE Conference on Computer Vision and
  Pattern Recognition}, 2019, pp. 821--830.

\bibitem{Nie_2019_ICCV}
Jing Nie, Rao~Muhammad Anwer, Hisham Cholakkal, Fahad~Shahbaz Khan, Yanwei
  Pang, and Ling Shao,
\newblock ``Enriched feature guided refinement network for object detection,''
\newblock in {\em The IEEE International Conference on Computer Vision (ICCV)},
  October 2019.

\bibitem{Duan_2019_ICCV}
Kaiwen Duan, Song Bai, Lingxi Xie, Honggang Qi, Qingming Huang, and Qi~Tian,
\newblock ``Centernet: Keypoint triplets for object detection,''
\newblock in {\em The IEEE International Conference on Computer Vision (ICCV)},
  October 2019.

\bibitem{lin2014microsoft}
Tsung-Yi Lin, Michael Maire, Serge Belongie, James Hays, Pietro Perona, Deva
  Ramanan, Piotr Doll{\'a}r, and C~Lawrence Zitnick,
\newblock ``Microsoft coco: Common objects in context,''
\newblock in {\em European conference on computer vision}. Springer, 2014, pp.
  740--755.

\bibitem{krishnavisualgenome}
Ranjay Krishna, Yuke Zhu, Oliver Groth, Justin Johnson, Kenji Hata, Joshua
  Kravitz, Stephanie Chen, Yannis Kalantidis, Li-Jia Li, David~A Shamma,
  et~al.,
\newblock ``Visual genome: Connecting language and vision using crowdsourced
  dense image annotations,''
\newblock {\em International Journal of Computer Vision}, vol. 123, no. 1, pp.
  32--73, 2017.

\bibitem{Liang2017DeepVR}
Xiaodan Liang, Lisa Lee, and Eric~P. Xing,
\newblock ``Deep variation-structured reinforcement learning for visual
  relationship and attribute detection,''
\newblock {\em 2017 IEEE Conference on Computer Vision and Pattern Recognition
  (CVPR)}, pp. 4408--4417, 2017.

\bibitem{yi2018neural}
Kexin Yi, Jiajun Wu, Chuang Gan, Antonio Torralba, Pushmeet Kohli, and
  Joshua~B. Tenenbaum,
\newblock ``Neural-symbolic vqa: Disentangling reasoning from vision and
  language understanding,''
\newblock in {\em Advances in Neural Information Processing Systems}, 2018, pp.
  1039--1050.

\bibitem{densecap}
Justin Johnson, Andrej Karpathy, and Li~Fei-Fei,
\newblock ``Densecap: Fully convolutional localization networks for dense
  captioning,''
\newblock in {\em Proceedings of the IEEE Conference on Computer Vision and
  Pattern Recognition}, 2016.

\bibitem{anderson2018bottom}
Peter Anderson, Xiaodong He, Chris Buehler, Damien Teney, Mark Johnson, Stephen
  Gould, and Lei Zhang,
\newblock ``Bottom-up and top-down attention for image captioning and visual
  question answering,''
\newblock in {\em Proceedings of the IEEE conference on computer vision and
  pattern recognition}, 2018, pp. 6077--6086.

\bibitem{han2019grounding}
Xudong Han, Philip Schulz, and Trevor Cohn,
\newblock ``Grounding learning of modifier dynamics: An application to color
  naming,''
\newblock in {\em Proceedings of the 2019 Conference on Empirical Methods in
  Natural Language Processing and the 9th International Joint Conference on
  Natural Language Processing (EMNLP-IJCNLP)}, 2019, pp. 1488--1493.

\bibitem{winn2018lighter}
Olivia Winn and Smaranda Muresan,
\newblock ``‘lighter’can still be dark: Modeling comparative color
  descriptions,''
\newblock in {\em Proceedings of the 56th Annual Meeting of the Association for
  Computational Linguistics (Volume 2: Short Papers)}, 2018, pp. 790--795.

\bibitem{he2016deep}
Kaiming He, Xiangyu Zhang, Shaoqing Ren, and Jian Sun,
\newblock ``Deep residual learning for image recognition,''
\newblock in {\em Proceedings of the IEEE conference on computer vision and
  pattern recognition}, 2016, pp. 770--778.

\bibitem{Simonyan15}
Karen Simonyan and Andrew Zisserman,
\newblock ``Very deep convolutional networks for large-scale image
  recognition,''
\newblock in {\em International Conference on Learning Representations}, 2015.

\bibitem{girshick2015fast}
Ross Girshick,
\newblock ``Fast r-cnn,''
\newblock in {\em Proceedings of the IEEE international conference on computer
  vision}, 2015, pp. 1440--1448.

\bibitem{hudson2018gqa}
Drew~A Hudson and Christopher~D Manning,
\newblock ``Gqa: A new dataset for real-world visual reasoning and
  compositional question answering,''
\newblock {\em Conference on Computer Vision and Pattern Recognition (CVPR)},
  2019.

\bibitem{lin2017feature}
Tsung-Yi Lin, Piotr Doll{\'a}r, Ross Girshick, Kaiming He, Bharath Hariharan,
  and Serge Belongie,
\newblock ``Feature pyramid networks for object detection,''
\newblock in {\em Proceedings of the IEEE conference on computer vision and
  pattern recognition}, 2017, pp. 2117--2125.

\bibitem{kingma205adam}
Diederick~P Kingma and Jimmy Ba,
\newblock ``Adam: A method for stochastic optimization,''
\newblock in {\em International Conference on Learning Representations (ICLR)},
  2015.

\end{thebibliography}

\end{document}